# Conditional Independence in Uncertainty Theories

Prakash P. Shenoy
School of Business
University of Kansas
Summerfield Hall
Lawrence, KS 66045-2003 USA
pshenoy@ukanvm.bitnet

## Abstract

This paper introduces the notions of independence and conditional independence in valuation-based systems (VBS). VBS is an axiomatic framework capable of representing many different uncertainty calculi. We define independence and conditional independence in terms of factorization of the joint valuation. The definitions of independence and conditional independence in VBS generalize the corresponding definitions in probability theory. Our definitions apply not only to probability theory, but also to Dempster-Shafer's belief-function theory, Spohn's epistemic-belief theory, and Zadeh's possibility theory. In fact, they apply to any uncertainty calculi that fit in the framework of valuation-based systems.

## 1 INTRODUCTION

The concept of conditional independence between two subsets of variables given a third has been extensively studied in probability theory [Dawid 1979, Spohn 1980, Lauritzen 1989, Pearl 1988, Smith 1989, Geiger 1990]. The concept of conditional independence in probability theory has been interpreted in terms of relevance. If r, s and t are disjoint subsets of variables, then to say that r and s are conditionally independent given t, means that the conditional distribution of r, given values of s and t, are governed by the value of t alone—further information about the value of s is irrelevant.

The concept of conditional independence for variables has also been studied in Spohn's theory of epistemic beliefs [Spohn 1988, Hunter 1991]. However, the concept of independence for *variables* has not been studied in Dempster-Shafer's theory of belief functions [Dempster 1967, Shafer 1976] or in Zadeh's possibility theory [Zadeh 1979, Dubois and Prade 1988].[1]

---

[1] Dempster [1967], Shafer [1976, 1982, 1984, 1987, 1990], and Smets [1986] have defined independence for belief functions, but not for *variables* on which belief functions are defined. Shafer [1976] has defined independence for frames of discernment, a concept further studied by Shafer, Shenoy and Mellouli [1987]. Belief functions in belief-function theory are analogs of probability functions in probability theory.

An abstract framework that unifies various uncertainty calculi is that of valuation-based systems [Shenoy 1989, 1991a]. In VBS, knowledge about a set of variables is represented by a valuation for that set of variables. There are three operators in VBS that are used to make inferences. These are called combination, marginalization, and removal. Combination represents aggregation of knowledge. Marginalization represents coarsening of knowledge. And removal represents disaggregation of knowledge.

The framework of VBS is able to uniformly represent probability theory, Dempster-Shafer's belief-function theory, Spohn's epistemic-belief theory, and Zadeh's possibility theory. In this paper, we will develop the notion of independence and conditional independence for variables in the framework of VBS. One advantage of this generality is that all results developed here will apply uniformly to all uncertainty calculi that fit in the framework of VBS. Thus the results described in this paper apply to, for example, probability theory, Dempster-Shafer's belief-function theory, Spohn's epistemic-belief theory, and Zadeh's possibility theory.

What does it mean for two disjoint subsets of variables to be independent? Intuitively, independence can be defined in terms of factorization of the joint valuation. If τ is a valuation for r∪s, then we say that r and s are independent with respect to τ iff τ factors into two valuations, one whose domain only involves r, and the other whose domain only involves s. One implication of this is that if we are interested in constructing a valuation for r∪s, then independence of r and s allows us to construct this valuation by, first, constructing two valuations—one whose domain involving only r and the other whose domain involving only s—and second, by simply combining the two valuations to get the result.

What does it mean for two disjoint subsets of variables to be conditionally independent given a third disjoint subset? Conditional independence can also be described in terms of factorization of the joint valuation. Suppose τ is a valuation for r∪s∪t. We say r and s are conditionally independent given t with respect to τ iff the valuation τ factors into two valuations, one whose domain involves variables in r∪t, and the other whose domain involves only variables in s∪t.

An outline of this paper is as follows. In section 2, we describe the framework of valuation-based systems (VBS). The VBS framework was described earlier in [Shenoy 1989, 1991a]. Here we extend the framework by defining three new classes of valuations called normal, proper normal, and positive proper normal. Also, we introduce a new operator called removal, and define some new axioms for the removal operator. Shenoy [1991b] shows how probability theory, Dempster-Shafer's belief-function theory, Spohn's epistemic-belief theory, and Zadeh's possibility theory fit in the framework of VBS.

In section 3, we define independence and conditional independence for sets of variables. We show that these definitions satisfy some well known properties that have been stated by Dawid [1979], Spohn [1980], Lauritzen [1989], Pearl [1988], and Smith [1989] in the context of probability theory. Using Pearl's terminology, the conditional independence relation in VBS is a graphoid. Finally, in section 4, we make some concluding remarks. Proofs of all results can be found in [Shenoy 1991b].

## 2 VALUATION-BASED SYSTEMS

In this section, we describe the framework of valuation-based systems (VBS). In a VBS, we represent knowledge by entities called variables and valuations. We infer independence relations using three operators called combination, marginalization, and removal. We use these operators on valuations.

The framework of VBS is described in [Shenoy 1989, 1991a]. The motivation there was to describe a local computational method for computing marginals of the joint valuation. In this paper, we embellish the VBS framework by introducing three new classes of valuations called normal, proper normal, and positive proper normal, and by introducing a new operator called removal. Our motivation here is to define independence and describe its properties.

**Variables** We assume there is a finite set $\mathcal{X}$ whose elements are called *variables*. Variables will be denoted by upper-case letters, X, Y, Z, etc. Subsets of $\mathcal{X}$ will be denoted by lower-case letters, r, s, t, etc.

**Valuations** For each $s \subseteq \mathcal{X}$, there is a set $\mathcal{V}_s$. We call the elements of $\mathcal{V}_s$ *valuations for s*. Let $\mathcal{V}$ denote $\cup \{\mathcal{V}_s \mid s \subseteq \mathcal{X}\}$, the set of all *valuations*. If $\sigma$ is a valuation for s, then we say that s is the *domain* of $\sigma$. Valuations will be denoted by lower-case Greek alphabets, $\rho, \sigma, \tau$, etc.

Valuations are primitives in our abstract framework and, as such, require no definition. But as we shall see shortly, they are objects which can be combined, marginalized, and removed. A valuation for s represents some knowledge about variables in s.

In probability theory, with each variable X, we associate a set $\mathcal{W}_X$ called the *frame for X*. Also for each $s \subseteq \mathcal{X}$, we associate the set $\mathcal{W}_s = \times \{\mathcal{W}_X \mid X \in s\}$ called the *frame for s*. Elements of $\mathcal{W}_s$ are called *configurations of s*. In probability theory, for example, a valuation for s is a function $\sigma: \mathcal{W}_s \to \mathbb{R}$, where $\mathbb{R}$ is the set of all real numbers.

**Zero Valuations** For each $s \subseteq \mathcal{X}$, there is at most one valuation $\zeta_s \in \mathcal{V}_s$ called the *zero valuation for s*. Let $\mathcal{Z}$ denote $\{\zeta_s \mid s \subseteq \mathcal{X}\}$, the set of all *zero valuations*. Note that we are not assuming zero valuations always exist. If zero valuations do not exist, $\mathcal{Z} = \emptyset$. We call valuations in $\mathcal{V} - \mathcal{Z}$ *nonzero valuations*. Intuitively, a zero valuation represents knowledge that is internally inconsistent. In probability theory, for example, a zero valuation for s is the valuation $\zeta_s$ such that $\zeta_s(x) = 0$ for all $x \in \mathcal{W}_s$.

**Proper Valuations** For each subset s of $\mathcal{X}$, there is a subset $\mathcal{P}_s$ of $\mathcal{V}_s - \{\zeta_s\}$. We call the elements of $\mathcal{P}_s$ *proper valuations for s*. Let $\mathcal{P}$ denote $\cup \{\mathcal{P}_s \mid s \subseteq \mathcal{X}\}$, the set of all *proper valuations*. Intuitively, a proper valuation represents knowledge that is partially coherent. In probability theory, for example, a proper valuation is a nonzero valuation $\sigma$ such that $\sigma(x) \geq 0$ for all $x \in \mathcal{W}_s$.

**Normal Valuations** For each $s \subseteq \mathcal{X}$, there is another subset $\mathcal{N}_s$ of $\mathcal{V}_s - \{\zeta_s\}$. We call the elements of $\mathcal{N}_s$ *normal valuations for s*. Let $\mathcal{N}$ denote $\cup \{\mathcal{N}_s \mid s \subseteq \mathcal{X}\}$, the set of all *normal valuations*. Intuitively, a normal valuation represents knowledge that is partially coherent in a sense different from proper valuations. In probability theory, for example, a normal valuation is a nonzero valuation $\sigma$ such that $\Sigma \{\sigma(x) \mid x \in \mathcal{W}_s\} = 1$.

**Proper Normal Valuations** For each $s \subseteq \mathcal{X}$, let $\mathcal{R}_s$ denote $\mathcal{P}_s \cap \mathcal{N}_s$. We call the elements of $\mathcal{R}_s$ *proper normal valuations for s*. Let $\mathcal{R}$ denote $\cup \{\mathcal{R}_s \mid s \subseteq \mathcal{X}\}$, the set of all *proper normal valuations*. Intuitively, a proper normal valuation represents knowledge that is completely coherent.

**Positive Proper Normal Valuations** For each $s \subseteq \mathcal{X}$, there is a subset $\mathcal{R}_s^+$ of $\mathcal{R}_s$. We call the elements of $\mathcal{R}_s^+$ *positive proper normal valuations for s*. Let $\mathcal{R}^+$ denote $\cup \{\mathcal{R}_s^+ \mid s \subseteq \mathcal{X}\}$, the set of all *positive proper normal valuations*. As we will see later, positive proper normal valuations are proper normal valuations that have unique identities. In probability theory, for example, a positive proper normal valuation for s is a proper normal valuation $\sigma$ such that $\sigma(x) > 0$ for all $x \in \mathcal{W}_s$.

Figure 1 shows the relations between the different types of valuations. As per our definitions, $\mathcal{Z} \subseteq \mathcal{V}$, $\mathcal{P} \subseteq (\mathcal{V} - \mathcal{Z})$, $\mathcal{N} \subseteq (\mathcal{V} - \mathcal{Z})$, $\mathcal{R} = \mathcal{P} \cap \mathcal{N}$, and $\mathcal{R}^+ \subseteq \mathcal{R}$.

**Combination** We assume there is a mapping $\oplus: \mathcal{V} \times \mathcal{V} \to \mathcal{N} \cup \mathcal{Z}$, called *combination*, such that the following axioms are satisfied:

(C1) (*Domain*) If $\rho$ and $\sigma$ are valuations for r and s, respectively, then $\rho \oplus \sigma$ is a valuation for $r \cup s$.

(C2) (*Associative*) $\rho \oplus (\sigma \oplus \tau) = (\rho \oplus \sigma) \oplus \tau$.

(C3) (*Commutative*) $\rho \oplus \sigma = \sigma \oplus \rho$.

(C4) (*Zero*) Suppose zero valuations exist, suppose $\sigma$ is a valuation for s, and suppose $\rho$ is a valuation for r. Then $\sigma \oplus \zeta_r = \zeta_s \oplus \rho = \zeta_{r \cup s}$.

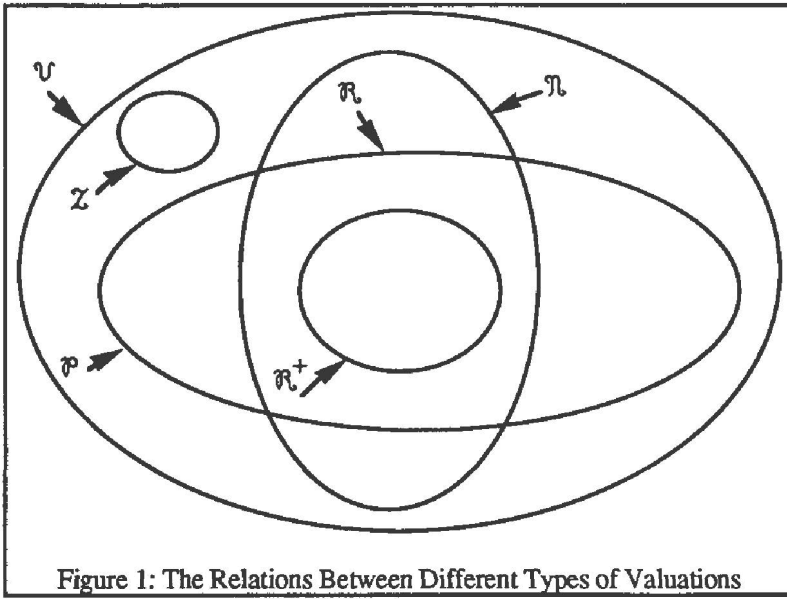

Figure 1: The Relations Between Different Types of Valuations

(C5) (*Nonzero*) If $\rho$ and $\sigma$ are both nonzero valuations, then $\rho \oplus \sigma$ is either normal or zero.

(C6) (*Proper*) If $\rho$ and $\sigma$ are both proper valuations, then $\rho \oplus \sigma$ is either proper normal or zero.

(C7) (*Positive Proper Normal Valuations*) If $\rho$ and $\sigma$ are both positive proper normal, then $\rho \oplus \sigma$ is positive proper normal.

If $\rho \oplus \sigma$, read as $\rho$ plus $\sigma$, is a zero valuation, then we say that $\rho$ and $\sigma$ are *inconsistent*. If $\rho \oplus \sigma$ is a normal valuation, then we say that $\rho$ and $\sigma$ are *consistent*.

Intuitively, combination corresponds to aggregation of knowledge. If $\rho$ and $\sigma$ are valuations for r and s representing knowledge about variables in r and s, respectively, then $\rho \oplus \sigma$ represents the aggregated knowledge about variables in r∪s. In probability theory, for example, combination is pointwise multiplication followed by normalization (if normalization is possible).

An implication of Axiom C2 is that when we have multiple combinations of valuations, we can write it without using parenthesis. For example, $(...((\sigma_1 \oplus \sigma_2) \oplus \sigma_3) \oplus ... \oplus \sigma_m)$ can be written simply as $\sigma_1 \oplus ... \oplus \sigma_m$ without parenthesis. Further, by Axiom C3, we can write $\sigma_1 \oplus ... \oplus \sigma_m$ simply as $\oplus \{\sigma_1, ..., \sigma_m\}$, i.e., not only do we not need parenthesis, we need not indicate the order in which the valuations are combined. Mathematically, Axioms C2 and C3 imply that the pair $(\mathcal{V}, \oplus)$ is a commutative semigroup [Petrich 1973].

An implication of Axioms C1, C2, C3, C4 and C5 is that the set $\mathfrak{N}_s \cup \{\zeta_s\}$ together with the combination operation $\oplus$ can be regarded as a commutative subsemigroup. (If zero valuations do not exist, then $\mathfrak{N}_s \cup \{\zeta_s\} = \mathfrak{N}_s$). By Axiom C4, if zero valuations exist, then the valuation $\zeta_s$ is the zero of the subsemigroup $\mathfrak{N}_s \cup \{\zeta_s\}$. It follows from Axiom C6 that for each $s \subseteq \mathfrak{X}$, $(\mathfrak{N}_s \cup \{\zeta_s\}, \oplus)$ is a commutative subsemigroup, and it follows from Axiom C7 that for each $s \subseteq \mathfrak{X}$, $(\mathfrak{R}_s^+, \oplus)$ is a commutative subsemigroup.

**Identity Valuations** We will assume that for each $s \subseteq \mathfrak{X}$, and for each $\sigma \in \mathfrak{N}_s \cup \{\zeta_s\}$, there exists at least one identity for it, i.e., there exists $\delta_\sigma \in \mathfrak{N}_s \cup \{\zeta_s\}$ such that $\sigma \oplus \delta_\sigma = \sigma$. A valuation may have more than one identity. Axiom C4 states that every element of $\mathfrak{N}_s \cup \{\zeta_s\}$ is an identity for $\zeta_s$. Note that if $\sigma \in \mathfrak{N}_s$, then $\delta_\sigma \in \mathfrak{N}_s$ (Proof: If $\delta_\sigma = \zeta_s$, then $\sigma \oplus \delta_\sigma = \sigma \oplus \zeta_s = \zeta_s \neq \sigma$, contradicting the fact that $\delta_\sigma$ is an identity for $\sigma$).

Also, we will assume that for each $s \subseteq \mathfrak{X}$, the commutative subsemigroup $\mathfrak{N}_s \cup \{\zeta_s\}$ has an identity, denoted by $\iota_s$, which is positive proper normal. In other words, there exists $\iota_s \in \mathfrak{R}_s^+$ such that for each $\sigma \in \mathfrak{N}_s$, $\sigma \oplus \iota_s = \sigma$. Note that a commutative semigroup may have at most one identity. Also, $\iota_s$ is an identity for each element of $\mathfrak{N}_s \cup \{\zeta_s\}$.

We will assume that for each $s \subseteq \mathfrak{X}$, and for each $\sigma \in \mathfrak{R}_s \cup \{\zeta_s\}$, there exists at least one identity for it, i.e., there exists $\delta_\sigma \in \mathfrak{R}_s \cup \{\zeta_s\}$ such that $\sigma \oplus \delta_\sigma = \sigma$. Note that if $\sigma \in \mathfrak{R}_s$, then $\delta_\sigma \in \mathfrak{R}_s$. Also, since $\mathfrak{R}_s$ is a subset of $\mathfrak{N}_s$, and $\iota_s \in \mathfrak{R}_s$, $\iota_s$ is also the identity for the semigroup $\mathfrak{R}_s \cup \{\zeta_s\}$.

We will assume that for each $s \subseteq \mathfrak{X}$, each element of $\mathfrak{R}_s^+$ has a unique identity. Since $\mathfrak{R}_s^+ \subseteq \mathfrak{R}_s$, and $\iota_s \in \mathfrak{N}_s^+$, this implies that $\iota_s$ is the identity for each $\sigma$ in $\mathfrak{R}_s^+$, i.e. $\delta_\sigma = \iota_s$ if $\sigma \in \mathfrak{R}^+$, and that $\iota_s$ is the identity for $\mathfrak{R}_s^+$.

In probability theory, for example, the identity $\iota_s$ for $\mathfrak{N}_s \cup \{\zeta_s\}$ is given by $\iota_s(x) = 1/|\mathcal{W}_s|$ for all $x \in \mathcal{W}_s$. Suppose $\sigma$ is a normal valuation for s. An identity $\delta_\sigma$ for $\sigma$ is a proper normal valuation for s such that $\delta_\sigma(x) = 1/K$ if $\sigma(x) \neq 0$, and $\delta_\sigma(x) =$ either 0 or 1/K if $\sigma(x) = 0$. K is a constant whose value is determined by the fact that $\delta_\sigma$ is a normal valuation.

**Valuations for the Empty Set** We will assume that the set $\mathfrak{N}_\emptyset$ consists of exactly one element. This assumption implies that $\mathfrak{R}_\emptyset^+ = \mathfrak{R}_\emptyset = \mathfrak{N}_\emptyset = \{\iota_\emptyset\}$ where $\iota_\emptyset$ is the identity valuation for the empty set. Also, we will assume that if $\sigma$ is nonzero valuation for s, and $\beta_\emptyset$ is a nonzero valuation for the empty set, then $\sigma \oplus \beta_\emptyset$ is not zero, i.e., $\sigma \oplus \beta_\emptyset$ is a normal valuation for s. This assumption implies that if $\alpha_\emptyset$ and $\beta_\emptyset$ are nonzero valuations for the empty set, then $\alpha_\emptyset \oplus \beta_\emptyset = \iota_\emptyset$.

If $\sigma$ is a nonzero valuation that is not normal, then $\sigma \oplus \iota_\emptyset \neq \sigma$ (since $\sigma \oplus \iota_\emptyset$ is normal and $\sigma$ is not). We can regard $\sigma \oplus \iota_\emptyset$ as the "normalized" form of $\sigma$.

**Marginalization** We assume that for each $s \subseteq \mathfrak{X}$, and for each $X \in s$, there is a mapping $\downarrow (s-\{X\}): \mathcal{V}_s \to \mathcal{V}_{s-\{X\}}$, called *marginalization to s–{X}*, such that it satisfies the following axioms:

(M1) (*Domain*) If $\sigma$ is a valuation for s, then $\sigma^{\downarrow(s-\{X\})}$ is a valuation for $s-\{X\}$.

(M2) (*Order of Deletion*) Suppose $\sigma$ is a valuation for s, and suppose $X_1, X_2 \in s$. Then $(\sigma^{\downarrow(s-\{X_1\})})^{\downarrow(s-\{X_1,X_2\})}$ $= (\sigma^{\downarrow(s-\{X_2\})})^{\downarrow(s-\{X_1,X_2\})}$.

(M3) (*Nonzero*) $\sigma^{\downarrow(s-\{X\})}$ is nonzero iff $\sigma$ is nonzero.

(M4) (*Proper*) If $\sigma$ is a proper valuation, then $\sigma^{\downarrow(s-\{X\})}$ is a proper valuation.

(M5) (*Normal*) $\sigma^{\downarrow(s-\{X\})}$ is normal iff $\sigma$ is normal.

(M6) (*Positive Proper Normal*) If $\sigma$ is positive proper normal, then $\sigma^{\downarrow(s-\{X\})}$ is positive proper normal.

We call $\sigma^{\downarrow(s-\{X\})}$ the *marginal of $\sigma$ for $s-\{X\}$*.

Intuitively, marginalization corresponds to coarsening of knowledge. If $\sigma$ is a valuation for s representing some knowledge about variables in s, and $X \in s$, then $\sigma^{\downarrow(s-\{X\})}$ represents the knowledge about variables in $s-\{X\}$ implied by $\sigma$ if we disregard variable X. In probability theory, for example, marginalization to $s-\{X\}$ is addition over the frame for X.

*If we regard marginalization as a coarsening of a valuation by deleting variables, then Axiom M2 says that the order in which the variables are deleted does not matter.*[2] One implication of this axiom is that $(\sigma^{\downarrow(s-\{X_1\})})^{\downarrow(s-\{X_1,X_2\})}$ can be written simply as $\sigma^{\downarrow(s-\{X_1,X_2\})}$, i.e., we need not indicate the order in which the variables are deleted.

Axiom M3 is vacuous if zero valuations do not exist. An implication of M3 is that $\zeta_s^{\downarrow(s-\{X\})} = \zeta_{s-\{X\}}$. An implication of Axiom M5 is that a valuation $\sigma$ for s is normal iff $\sigma^{\downarrow\emptyset} = \iota_\emptyset$.

We will assume further that the marginalization and combination operators satisfies the following two axioms.

**Axiom CM1** (*Combination and Marginalization 1*) Suppose $\sigma$ is a normal valuation for s, suppose $r \subseteq s$, and suppose $\delta_{\sigma\downarrow r}$ is an identity for $\sigma^{\downarrow r}$. Then

$$\sigma \oplus \delta_{\sigma\downarrow r} = \sigma.$$

**Axiom CM2** (*Combination and Marginalization 2*) Suppose $\rho$ and $\sigma$ are valuations for r and s, respectively. Suppose $X \notin r$, and suppose $X \in s$. Then
$$(\rho \oplus \sigma)^{\downarrow((r \cup s)-\{X\})} = \rho \oplus (\sigma^{\downarrow(s-\{X\})}).$$

The following lemma states some easy implications of Axiom CM1.[3]

**Lemma 2.1** Suppose Axioms C1–C7, M1–M6, and CM1 hold. Then the following statements hold.

(i). If $\sigma$ is a normal valuation for s, and $r \subseteq s$, then $\sigma \oplus \iota_r = \sigma$.

(ii). If $\sigma$ and $\rho$ are nonzero valuations, then $\sigma \oplus \rho^{\downarrow\emptyset} = \sigma \oplus \iota_\emptyset$.

(iii). $\sigma$ is normal or zero iff $\sigma \oplus \iota_\emptyset = \sigma$.

(iv). If $r \subseteq s$, then $\iota_s \oplus \iota_r = \iota_s$.

(v). $\iota_s \oplus \iota_r = \iota_{r \cup s}$.

Axiom CM2 states that the computation of $(\rho \oplus \sigma)^{\downarrow((r \cup s)-\{X\})}$ can be accomplished without having to compute $\rho \oplus \sigma$. The combination $\rho \oplus \sigma$ is a valuation for $r \cup s$ whereas the combination $\rho \oplus (\sigma^{\downarrow(s-\{X\})})$ is a valuation for $(r \cup s)-\{X\}$. The following lemma is an easy consequence of Axiom CM2.[4]

**Lemma 2.2** Suppose Axioms C1–C3, M1, M2, and CM2 hold. Suppose $\rho$ is a valuation for r and suppose $\sigma$ is a valuation for s. Then $(\rho \oplus \sigma)^{\downarrow r} = \rho \oplus \sigma^{\downarrow r \cap s}$.

Axioms C1, C2, C3, M1, M2, and CM2 make local computation of marginals possible. Suppose $\{\sigma_1, ..., \sigma_m\}$ is a collection of valuations, and suppose $\sigma_i$ is a valuation for $s_i$. Suppose $\mathcal{X} = s_1 \cup ... \cup s_m$, and suppose $X \in \mathcal{X}$. Suppose we wish to compute $(\sigma_1 \oplus ... \oplus \sigma_m)^{\downarrow\{X\}}$. We can do so by successively deleting all variables but X from the collection of valuation $\{\sigma_1, ..., \sigma_m\}$. Each time we delete a variable, we do a fusion operation defined as follows. Consider a set of k valuations $\rho_1, ..., \rho_k$. Suppose $\rho_i$ is a valuation for $r_i$. Let $\text{Fus}_Y\{\rho_1, ..., \rho_k\}$ denote the collection of valuations after fusing the valuations in the set $\{\rho_1, ..., \rho_k\}$ with respect to variable $Y \in r_1 \cup ... \cup r_k$. Then

$$\text{Fus}_Y\{\rho_1, ..., \rho_k\} = \{\rho^{\downarrow(r-\{Y\})}\} \cup \{\rho_i \mid Y \notin r_i\}$$

where $\rho = \oplus\{\rho_i \mid Y \in r_i\}$, and $r = \cup\{r_i \mid Y \in r_i\}$. After fusion, the set of valuations is changed as follows. All valuations whose domains include Y are combined, and the resulting valuation is marginalized such that Y is eliminated from its domain. The valuations whose domains do not include Y remain unchanged. The following theorem describes the fusion algorithm, a method for computing $(\sigma_1 \oplus ... \oplus \sigma_m)^{\downarrow\{X\}}$ using only local computations.

---

[2] Axiom M2 is equivalent to the "consonance of marginalization" axiom in [Shenoy and Shafer 1990], which is stated as follows: If $\sigma$ is a valuation for s, and $q \subseteq r \subseteq s$, then $(\sigma^{\downarrow r})^{\downarrow q} = \sigma^{\downarrow q}$.

[3] In [Cano, Delgado, and Moral 1991, and Shafer 1991], statement (iv) of Lemma 2.1 is stated as an axiom. [Shafer 1991] proves statement (v) of Lemma 2.1 assuming statement (iv).

[4] The statement of Lemma 2.2 was first stated as an axiom in [Shenoy and Shafer 1990]. Shenoy [1991a] stated axiom CM2, which is stronger than the statement of Lemma 2.2. The added strength of axiom CM2 has advantages in the computation of marginals—see Theorem 2.1.

**Theorem 2.1** [Shenoy 1991a] Suppose $\{\sigma_1, ..., \sigma_m\}$ is a collection of valuations such that $\sigma_i$ is a valuation for $s_i$. Suppose Axioms C1, C2, C3, M1, M2, and CM2 hold. Let $\mathcal{X}$ denote $s_1 \cup ... \cup s_m$. Suppose $X \in \mathcal{X}$, and suppose $X_1 X_2 ... X_{n-1}$ is a sequence of variables in $\mathcal{X} - \{X\}$. Then

$$(\sigma_1 \oplus ... \oplus \sigma_m)^{\downarrow\{X\}} = \oplus \left\{ \mathrm{Fus}_{X_{n-1}} \{ ... \mathrm{Fus}_{X_2} \{ \mathrm{Fus}_{X_1} \{\sigma_1, ..., \sigma_m\} \} \} \right\}.$$

**Removal** We assume there is a mapping $\otimes: \mathcal{V} \times \mathcal{V} \to \mathcal{N} \cup \mathcal{Z}$, called *removal*, such that the following axioms are satisfied:

(R1) (*Domain*) Suppose $\sigma$ is a valuation for s, and suppose $\rho$ is a valuation for r. Then $\sigma \otimes \rho$ is a valuation for $r \cup s$;

(R2) (*Zero*) Suppose zero valuations exist, suppose $\sigma$ is a valuation for s, and suppose $\rho$ is a valuation for r. Then $\sigma \otimes \zeta_r = \zeta_s \otimes \rho = \zeta_{r \cup s}$.

(R3) (*Nonzero*) Suppose $\sigma$ and $\rho$ are nonzero valuations for s and r respectively. Then $\sigma \otimes \rho$ is either normal or zero.

(R4) (*Normal*) Suppose $\rho$ is a normal valuation for r. Then there exists an identity $\delta_\rho$ for $\rho$ such that $\rho \otimes \rho = \delta_\rho$.

We call $\sigma \otimes \rho$, read as $\sigma$ minus $\rho$, the *valuation resulting after removing $\rho$ from $\sigma$*.

Intuitively, $\sigma \otimes \rho$ can be interpreted as follows. If $\sigma$ and $\rho$ represent some knowledge, and if we remove the knowledge represented by $\rho$ from $\sigma$, then $\sigma \otimes \rho$ describes the knowledge that remains. In probability theory, for example, removal is pointwise division followed by normalization (if normalization is possible).

We assume the following two axioms that relate the removal operator to the combination and the marginalization operators.

**Axiom CR** (*Combination and Removal*) Suppose $\pi, \theta, \rho$ are valuations for p, q, and r, respectively. Then
$$(\pi \oplus \theta) \otimes \rho = \pi \oplus (\theta \otimes \rho),$$
$$\pi \otimes (\theta \oplus \rho) = (\pi \otimes \theta) \otimes \rho, \text{ and}$$
$$\pi \otimes (\theta \otimes \rho) = (\pi \otimes \theta) \oplus \rho$$

**Axiom MR** (*Marginalization and Removal*) Suppose $\sigma$ is a valuation for s, suppose $\rho$ is a valuation for r, suppose $X \in s$, and suppose $X \notin r$. Then
$$(\sigma \otimes \rho)^{\downarrow((r \cup s) - \{X\})} = \sigma^{\downarrow(s - \{X\})} \otimes \rho.$$

It follows from Axioms C2, C3 and CR that $(\pi \oplus \theta) \otimes \rho = (\pi \otimes \rho) \oplus \theta$. The following lemma states some easy consequences of Axiom R4.

**Lemma 2.3** Suppose $\sigma$ is a valuation for s, and suppose $\rho$ is a normal valuation for r. Then the following statements hold.

(i). $((\sigma \oplus \rho) \otimes \rho)^{\downarrow s} = \sigma \oplus \iota_\emptyset$.

(ii). If $\sigma$ is normal, then $((\sigma \oplus \rho) \otimes \rho)^{\downarrow s} = \sigma$.

(iii). $((\sigma \oplus \rho) \otimes \rho)^{\downarrow s} \oplus \rho = \sigma \oplus \rho$.

(iv). $((\sigma \oplus \rho) \otimes \rho) \oplus \rho = \sigma \oplus \rho$.

(v). If $\rho$ is positive proper normal, then $\rho \otimes \rho = \iota_r$.

(vi). If $\rho$ is positive proper normal, then $(\sigma \oplus \rho) \otimes \rho = \sigma \oplus \iota_r$.

(vii). If $\sigma$ is normal, and $\rho$ is positive proper normal, then $(\sigma \oplus \rho) \otimes \rho = \sigma$.

(viii). If $\sigma$ is normal, and $r \subseteq s$, then $(\sigma \otimes \sigma^{\downarrow r}) \oplus \sigma^{\downarrow r} = \sigma$.

(ix). If $\sigma$ is normal, and $r \subseteq s$, then $\sigma \otimes \sigma^{\downarrow r}$ is normal.

(x). If $\sigma$ is normal, and $r \subseteq s$, then $\sigma \otimes \iota_r = \sigma$.

(xi). If $\sigma$ is normal, and $r \subseteq s$, then there exists an identity $\delta_{\sigma^{\downarrow r}}$ for $\sigma^{\downarrow r}$ such that $(\sigma \otimes \sigma^{\downarrow r})^{\downarrow r} = \delta_{\sigma^{\downarrow r}}$.

**Conditional Valuations.** Suppose $\sigma$ is a proper normal valuation for s, and suppose $r \subseteq s$. The normal valuation $\sigma \otimes \sigma^{\downarrow r}$ for s plays an important role in the theory of independence. Borrowing terminology from probability theory, we call $\sigma \otimes \sigma^{\downarrow r}$ the *conditional for s–r given r*. Conditional valuations have two important properties: $(\sigma \otimes \sigma^{\downarrow r}) \oplus \sigma^{\downarrow r} = \sigma$, and $(\sigma \otimes \sigma^{\downarrow r})^{\downarrow r} = \delta_{\sigma^{\downarrow r}}$.

## 3 INDEPENDENCE AND CONDITIONAL INDEPENDENCE

In this section, we define independence and conditional independence in terms of factorization of the joint valuation. Also, we show that these definitions imply the well known properties of independence and conditional independence in probability theory [Dawid 1979, Spohn 1980, Lauritzen 1989] and in other domains [Pearl 1988, Smith 1989].

The essence of independence is as follows. We say disjoint subsets r and s are independent with respect to a proper normal valuation $\tau$ iff $\tau^{\downarrow(r \cup s)}$ factors into two valuations $\rho$ and $\sigma$, where $\rho$ is a valuation for r, and $\sigma$ is a valuation for s.

The definition of independence is either objective or subjective depending on whether we have an objective or subjective measure of knowledge represented by proper normal valuation $\tau$. In probability theory, in some cases, we start with an objective specification of a joint probability distribution of all variables. This joint probability distribution then serves as an objective measure of knowledge, and all statements of independence are objective with respect to this state of knowledge. In other cases, however, we do not start always with a joint probability distribution. In such cases, the first task is to specify a joint probability distribution. To make this specification task simpler, we make assertions of independence that are necessarily subjective. However, once we have a specification of a joint probability distribution (obtained either objec-tively or subjectively), all further statements of independence are necessarily objective with respect to the joint probability distribution.

Let $\tau$ be a proper normal valuation for $\mathcal{X}$. We will henceforth assume that $\tau$ represents the global knowledge regarding all variables in the VBS. For example, in probability theory, $\tau$ would represent the joint probability distribution for all variables in $\mathcal{X}$.

**Definition 3.1** (*Independence*) Suppose $\tau$ is a proper normal valuation for $\mathcal{X}$, and suppose r, s $\subseteq \mathcal{X}$, r$\cap$s = $\emptyset$. We say *r and s are independent with respect to $\tau$*, written as r $\perp_\tau$ s, iff $\tau^{\downarrow(r\cup s)} = \rho \oplus \sigma$, where $\rho$ and $\sigma$ are valuations for r and s, respectively.

When it is clear that all independence statements are with respect to $\tau$, we will simply say 'r and s are independent' instead of 'r and s are independent with respect to $\tau$,' and use the simpler notation r $\perp$ s instead of r $\perp_\tau$ s.

**Theorem 3.1** (*Symmetry*) Suppose $\tau$ is a proper normal valuation for $\mathcal{X}$, and suppose r, s $\subseteq \mathcal{X}$, r$\cap$s = $\emptyset$. If r $\perp$ s, then s $\perp$ r.

The following lemma gives alternative characterizations of the independence relation.[5]

**Lemma 3.1** Suppose $\tau$ is a proper normal valuation for $\mathcal{X}$, and suppose r, s $\subseteq \mathcal{X}$, r$\cap$s = $\emptyset$. The following statements are equivalent.

(i). r $\perp$ s.

(ii). $\tau^{\downarrow(r\cup s)} = \rho \oplus \sigma$, where $\rho$ and $\sigma$ are valuations for r and s, respectively.

(iii). $\tau^{\downarrow(r\cup s)} = \tau^{\downarrow r} \oplus \tau^{\downarrow s}$.

(iv). There exists an identity $\delta_{\tau^{\downarrow r}}$ for $\tau^{\downarrow r}$ such that $\tau^{\downarrow(r\cup s)} \otimes \tau^{\downarrow r} = \sigma \oplus \delta_{\tau^{\downarrow r}}$, where $\sigma$ is a valuation for s.

(v). There exists an identity $\delta_{\tau^{\downarrow r}}$ for $\tau^{\downarrow r}$ such that $\tau^{\downarrow(r\cup s)} \otimes \tau^{\downarrow r} = \tau^{\downarrow s} \oplus \delta_{\tau^{\downarrow r}}$.

Definition 3.2 generalizes Definition 3.1 for any number of subsets of variables.

**Definition 3.2** (*Joint Independence*) Suppose $\tau$ is a proper normal valuation for $\mathcal{X}$, and suppose $r_1, ..., r_n$ are disjoint subsets of $\mathcal{X}$. We say $r_1, ..., r_n$ are *(jointly) independent with respect to $\tau$*, written as $\perp_\tau\{r_1, ..., r_n\}$, iff $\tau^{\downarrow(r_1\cup...\cup r_n)} = \rho_1 \oplus ... \oplus \rho_n$, where $\rho_i$ is a valuation for $r_i$, i = 1, ..., n.

---

[5] The statements of Lemma 3.1 are analogs of corresponding statements in [Dawid 1979] in the context of probability theory. Our contribution here is in showing that these statements hold in our more general framework of VBS. Thus they hold not only in probability theory (as shown by Dawid [1979]) but also in other uncertainty calculi that fit in the framework of VBS.

Definition 3.2 is a generalization of Definition 3.1. Note that r $\perp$ s iff $\perp\{r, s\}$. We know from probability theory that functions of independent random variables are independent. If $X_1$ and $X_2$ are independent random variables, then $f(X_1)$ and $g(X_2)$ are also independent random variables. More generally, if $X_1, ..., X_n$ are independent, $\{N_1, ..., N_k\}$ is a partition of the set $\{X_1, ..., X_n\}$, and $Y_j$ is a function of the $X_i$ in $N_j$, then $Y_1, ..., Y_k$ are independent. The following lemma makes an analogous statement.[6]

**Lemma 3.2** Suppose $\tau$ is a proper normal valuation for $\mathcal{X}$, and suppose $r_1, ..., r_n$ are disjoint subsets of $\mathcal{X}$. Suppose $\perp\{r_1, ..., r_n\}$. Suppose $\{N_1, ..., N_k\}$ is a partition of $\{1, ..., n\}$, i.e., $N_i \cap N_j = \emptyset$ if i $\neq$ j, and $N_1 \cup ... \cup N_k = \{1, ..., n\}$. Suppose $s_j \subseteq (\cup\{r_i | i \in N_j\})$, for j = 1, ..., k. Then $\perp\{s_1, ..., s_k\}$.

The statement in the following corollary to Lemma 3.2 is called decomposition [Pearl 1988]. It is a special case of Lemma 3.2.

**Corollary** (*Decomposition*) Suppose $\tau$ is a proper normal valuation for $\mathcal{X}$, suppose r, s, t are disjoint subsets of $\mathcal{X}$, and suppose r $\perp$ (s$\cup$t). Then r $\perp$ s.

The following lemma gives four alternative characterizations of joint independence.[7]

**Lemma 3.3** Suppose $\tau$ is a proper normal valuation for $\mathcal{X}$, and suppose $r_1, ..., r_n$ are disjoint subsets of $\mathcal{X}$. Then the following statements are equivalent.

(i). $\perp\{r_1, ..., r_n\}$

(ii). $\tau^{\downarrow(r_1\cup...\cup r_n)} = \rho_1 \oplus ... \oplus \rho_n$, where $\rho_i$ is a valuation for $r_i$, i = 1, ..., n.

(iii). $\tau^{\downarrow(r_1\cup...\cup r_n)} = \tau^{\downarrow r_1} \oplus ... \oplus \tau^{\downarrow r_n}$

(iv). $\perp\{r_1, ..., r_{n-1}\}$ and $(r_1\cup...\cup r_{n-1}) \perp r_n$.

(v). $r_i \perp \cup\{r_j | j = 1, ..., n, j \neq i\}$ for i = 1, ..., n.

(vi). $r_j \perp (r_1\cup...\cup r_{j-1})$ for j = 2, ..., n.

Definition 3.3 defines conditional independence for two subsets given a third.

**Definition 3.3** (*Conditional Independence*) Suppose $\tau$ is a proper normal valuation for $\mathcal{X}$, and suppose r, s, and t are disjoint subsets of $\mathcal{X}$. We say *r and s are conditionally independent given t with respect to $\tau$*, written as r $\perp_\tau$ s | t, iff $\tau^{\downarrow(r\cup s\cup t)} = \alpha_{r\cup t} \oplus \alpha_{s\cup t}$, where $\alpha_{r\cup t}$ and $\alpha_{s\cup t}$ are valuations for r$\cup$t and s$\cup$t, respectively.

The following lemma gives six alternative characterizations of conditional independence.

---

[6] An analogous statement is stated and proved in [Shafer, Shenoy, and Mellouli 1987] in the context of qualitative independence.

[7] The statements in Lemma 3.3 are analogs of corresponding statements in Shafer, Shenoy and Mellouli [1987] in the context of qualitative independence.

**Lemma 3.4** Suppose $\tau$ is a proper normal valuation for $\mathfrak{X}$, and suppose r, s, and t are disjoint subsets of $\mathfrak{X}$. The following statements are equivalent.

(i). $r \perp s \mid t$.

(ii). $\tau^{\downarrow(r \cup s \cup t)} = \alpha_{r \cup t} \oplus \alpha_{s \cup t}$, where $\alpha_{r \cup t}$ and $\alpha_{s \cup t}$ are valuations for $r \cup t$ and $s \cup t$, respectively.

(iii). $\tau^{\downarrow(r \cup s \cup t)} \otimes \tau^{\downarrow t} = \beta_{r \cup t} \oplus \beta_{s \cup t}$, where $\beta_{r \cup t}$ and $\beta_{s \cup t}$ are valuations for $r \cup t$ and $s \cup t$, respectively.

(iv). $\tau^{\downarrow(r \cup s \cup t)} =$
$\tau^{\downarrow t} \oplus (\tau^{\downarrow(r \cup t)} \otimes \tau^{\downarrow t}) \oplus (\tau^{\downarrow(s \cup t)} \otimes \tau^{\downarrow t})$.

(v). $\tau^{\downarrow(r \cup s \cup t)} \otimes \tau^{\downarrow t} =$
$(\tau^{\downarrow(r \cup t)} \otimes \tau^{\downarrow t}) \oplus (\tau^{\downarrow(s \cup t)} \otimes \tau^{\downarrow t})$.

(vi). $\tau^{\downarrow(r \cup s \cup t)} = (\tau^{\downarrow(r \cup t)} \otimes \tau^{\downarrow t}) \oplus \tau^{\downarrow(s \cup t)}$

(vii). There exists an identity $\delta_{\tau^{\downarrow(s \cup t)}}$ for $\tau^{\downarrow(s \cup t)}$ such that $\tau^{\downarrow(r \cup s \cup t)} \otimes \tau^{\downarrow(s \cup t)} = (\tau^{\downarrow(r \cup t)} \otimes \tau^{\downarrow t}) \oplus \delta_{\tau^{\downarrow(s \cup t)}}$.

(viii). There exists an identity $\delta_{\tau^{\downarrow(s \cup t)}}$ for $\tau^{\downarrow(s \cup t)}$ such that $\tau^{\downarrow(r \cup s \cup t)} \otimes \tau^{\downarrow(s \cup t)} = \alpha_{r \cup t} \oplus \delta_{\tau^{\downarrow(s \cup t)}}$.

Theorem 3.2 states another property of independence. This property is called weak union [Pearl 1988].

**Theorem 3.2** (*Weak Union*) Suppose $\tau$ is a proper normal valuation for $\mathfrak{X}$, and suppose r, s, and t are disjoint subsets of $\mathfrak{X}$. If $r \perp s \cup t$, then $r \perp s \mid t$.

Theorem 3.3 states another property of conditional independence. This property is called contraction [Pearl 1988].

**Theorem 3.3** (*Contraction*) Suppose $\tau$ is a proper normal valuation for $\mathfrak{X}$, and suppose r, s, and t are disjoint subsets of $\mathfrak{X}$. If $r \perp s$, and $r \perp t \mid s$, then $r \perp s \cup t$.

The next theorem states a property of conditional independence that holds only if the joint valuation $\tau$ is positive proper normal.

**Theorem 3.4** (*Intersection*) Suppose $\tau$ is a positive proper normal valuation for $\mathfrak{X}$, and suppose r, s, and t are disjoint subset of $\mathfrak{X}$. If $r \perp s \mid t$, and $r \perp t \mid s$, then $r \perp s \cup t$.

Definition 3.4 generalizes Definition 3.3 from two subsets to any number of subsets.

**Definition 3.4** (*Joint Conditional Independence*) Suppose $\tau$ is a proper normal valuation for $\mathfrak{X}$, and suppose $r_1, ..., r_n$, t are disjoint subsets of $\mathfrak{X}$. We say $r_1, ..., r_n$ *are conditionally independent given t with respect to* $\tau$, written as $\perp_\tau \{r_1, ..., r_k\} \mid t$, iff
$\tau^{\downarrow(r_1 \cup ... \cup r_n \cup t)} = \alpha_{r_1 \cup t} \oplus ... \oplus \alpha_{r_n \cup t}$, where $\alpha_{r_i \cup t}$ is a valuation for $r_i \cup t$, i = 1, ..., n.

Pearl and Paz [1987] call a conditional independence relation that satisfies symmetry, decomposition, weak union, contraction, and intersection *a graphoid*. From Theorems 3.1–3.4 and the corollary to Lemma 3.2, it follows that the definition of conditional independence in Definition 3.3 is a graphoid.

## 4 CONCLUSION

The main objective of this paper is to define independence and conditional independence in the framework of valuation-based systems. Although these concepts have been defined and extensively studied in probability theory, they have not been extensively studied in non-probabilistic uncertainty theories.

Drawing upon the literature on independence in probability theory [Dawid 1979, Spohn 1980, Lauritzen 1989, Pearl 1988, Smith 1989], we define independence and conditional independence in VBS. The framework of VBS was defined earlier by Shenoy [1989, 1991a]. However, the VBS framework defined there is inadequate for the purposes of studying properties of independence. In this paper, we embellish the framework by including three new classes of valuations called proper, normal, and positive proper normal, and by including a new operator called removal. The new definitions are stated in the form of axioms. Shenoy [1991b] shows that these axioms are general enough to include probability theory, Dempster-Shafer's belief-function theory, Spohn's epistemic belief theory, and Zadeh's possibility theory.

The framework of VBS as described in this paper enables us to define independence and conditional independence, and enables us to derive all major properties of conditional independence that have been derived in probability theory. Independence and conditional independence are defined in terms of factorization of the joint valuation. Thus, not only do we have a deeper understanding of independence in probability theory, we also understand what independence means in various non-probabilistic uncertainty theories. This should deflect some criticism that non-probabilistic uncertainty theories are not as well developed as probability theory.

### Acknowledgments

This work was supported in part by the National Science Foundation under grant IRI-8902444. I am grateful to Pierre Ndilikilikesha, Glenn Shafer, Philippe Smets, Leen-Kiat Soh, and an anonymous referee for comments and discussions.